\def\year{2018}\relax
\documentclass[letterpaper]{article} 
\usepackage{aaai19}  
\usepackage{times}  
\usepackage{helvet}  
\usepackage{courier}  
\usepackage{url}  
\usepackage{graphicx}  
\usepackage{booktabs}  
\usepackage{enumitem}
\usepackage{amsmath}
\usepackage{pdfpages} 
\usepackage{wrapfig}
\usepackage{multirow}
\usepackage{multicol}
\usepackage[skip=0pt]{subcaption}
\usepackage{footnote}
\usepackage{amssymb}
\usepackage{pifont}
\usepackage{bm}
\usepackage{commath}
\usepackage{tikz}
\usepackage{colortbl}
\usepackage{booktabs}
\usepackage{mathtools}
\usepackage{pdfpages}

\newcommand{\cmark}{\ding{51}}%
\newcommand{\xmark}{\ding{53}}%

\newcommand{\experterror}[0]{\varepsilon_{X\setminus\{i\}}}
\newcommand{\allerror}[0]{\varepsilon_{X}}
\newcommand\inlineeqno{\stepcounter{equation}\ (eq. \theequation)}

\begin{document}
%
\title{Granger-causal Attentive Mixtures of Experts:\\ Learning Important Features with Neural Networks}
\author{
  Patrick Schwab$^{1}$, Djordje Miladinovic$^{2}$, Walter Karlen$^{1}$ \\
  $^{1}$Institute of Robotics and Intelligent Systems, $^{2}$Department of Computer Science\\
  ETH Zurich, Switzerland\\
  \texttt{patrick.schwab@hest.ethz.ch} \\
}

\maketitle
\begin{abstract} 
Knowledge of the importance of input features towards decisions made by machine-learning models is essential to increase our understanding of both the models and the underlying data. Here, we present a new approach to estimating feature importance with neural networks based on the idea of distributing the features of interest among experts in an attentive mixture of experts (AME). AMEs use attentive gating networks trained with a Granger-causal objective to learn to jointly produce accurate predictions as well as estimates of feature importance in a single model. Our experiments show (i) that the feature importance estimates provided by AMEs compare favourably to those provided by state-of-the-art methods, (ii) that AMEs are significantly faster at estimating feature importance than existing methods, and (iii) that the associations discovered by AMEs are consistent with those reported by domain experts.
\end{abstract} 
\section{Introduction}
Neural networks are often criticised for being black-box models \cite{castelvecchi2016can}. Researchers have addressed this criticism by developing tools that provide visualisations and  explanations for the decisions of neural networks \cite{baehrens2010explain,simonyan2013deep,zeiler2014visualizing,xu2015show,shrikumar2016not,krause2016interacting,montavon2017explaining,koh2017understanding}. These explanations are desirable for the many machine-learning use cases in which both predictive performance and interpretability are of paramount importance \cite{kindermans2017reliability,smilkov2017smoothgrad,doshi2017towards}. They enable us to argue for the decisions of machine-learning models, show when algorithmic decisions might be biased or discriminating \cite{hardt2016equality}, help uncover the basis of decisions when there are legal or ethical circumstances that call for explanations \cite{goodman2016european}, and may facilitate the discovery of patterns that could advance our understanding of the underlying phenomena \cite{shrikumar2016not}.

\begin{figure}[t!]
\begin{center}
  \centerline{\includegraphics[width=\columnwidth]{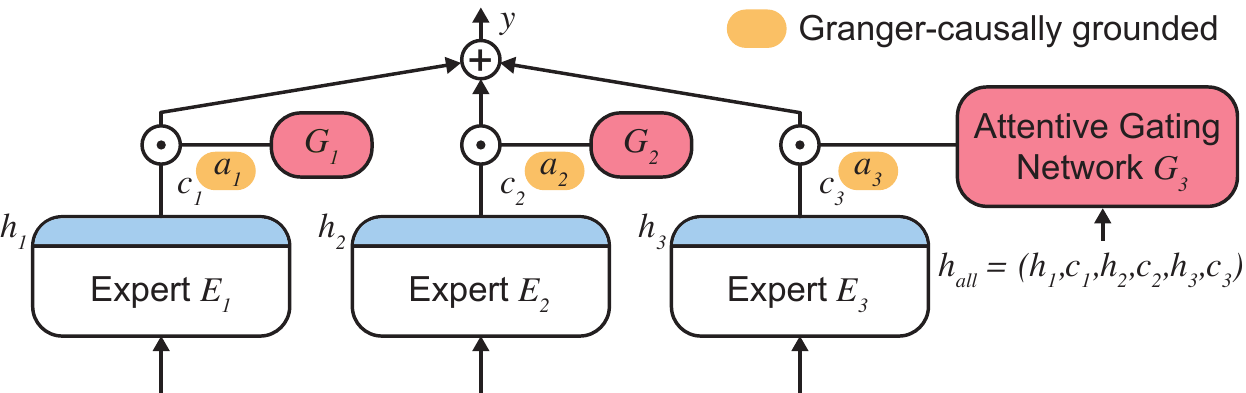}}
  \caption{An overview of attentive mixtures of experts (AMEs). The attentive gating networks $G_i$ (red) attend to the combined hidden state $h_\text{all}$ (blue) of the AME. Each expert's $G_i$ assigns an attentive factor $a_i$ to opportunistically control its contribution $c_i$ to the AME's final prediction $y$. }
 \label{fig:ame_overview}
\end{center}
\end{figure}

Estimating the relative contribution of individual input features towards outputs of a deep neural network is hard because the input features undergo multiple hierarchical, interdependent and non-linear transformations as they pass through the network \cite{montavon2017explaining}. We propose a new approach to feature importance estimation that optimises jointly for predictive performance and accurate assignment of feature importance in a single end-to-end trained neural network. Structurally, our approach builds on the idea of distributing the features of interest among experts in a mixture of experts \cite{jordan1994hierarchical}. The mixture of experts uses attentive gating networks to assign importance weights to individual experts (Figure \ref{fig:ame_overview}). However, when trained na\"ively, this structure alone does not generally learn to accurately assign weights that correspond to the importance of the experts' input features. We therefore draw upon a previously unreported connection between Granger-causality and feature importance estimation to define a secondary Granger-causal objective. Using the Granger-causal objective, we ensure that the weights given to individual experts correlate strongly and measurably with their ability to contribute to the decision at hand. Our experiments demonstrate that this optimisation-based approach towards learning to estimate feature importance leads to improvements of several orders of magnitude in computational performance over state-of-the-art methods. In addition, we show that the Granger-causal objective correlates with the expected quality of importance estimates, that AMEs compare favourably to the best existing methods in terms of feature importance estimation accuracy, and that AMEs discover associations that are consistent with those reported by domain experts\footnote{Source code at \url{https://github.com/d909b/ame}}.

{
\parfillskip=0pt
\parskip=0pt
\par} \begin{table*}[t]
\centering
\begin{small}
\begin{tabular}{l@{\hskip 14ex}*{5}{r}}
\toprule
 & AME & Attention & DeepLIFT & LIME & SHAP \\
 \midrule
Model-agnostic & \xmark & \xmark & \xmark & \cmark & \cmark \\
Measure of expected quality & \cmark & \xmark & \xmark & \xmark & \xmark \\
Computation Time ($\text{AME}=1$x) & 1x & 1x & 2x & 100-1000x & $>$1000x\\
\bottomrule
\end{tabular}
\end{small}
\caption{Comparison of AMEs to several representative methods for feature importance estimation. }
\label{tb:comparison}
\end{table*}

\textbf{Contributions.} We present the following contributions:
\begin{itemize}[noitemsep]
\item[(i)] We delineate an end-to-end trained AME that uses attentive gating to assign weights to individual experts.
\item[(ii)] We introduce a Granger-causal objective that measures the degree to which assigned feature importances correlate with the predictive value of features towards individual decisions.
\item[(iii)] We compare AMEs to state-of-the-art importance estimation methods on three datasets. The experiments show that AMEs are significantly faster than existing methods, that AMEs compare favourably to existing methods in terms of attribution accuracy, and that the associations discovered by AMEs are consistent with human experts.
\end{itemize}
\section{Related Work}
There are four main categories of approaches to assessing feature importance in neural networks: 

\paragraph{Perturbation-based Approaches.} Perturbation-based approaches attempt to explain the sensitivity of a machine-learning model to changes in its inputs by modelling the impact of local perturbations \cite{ribeiro2016should,adler2016auditing,fong2017interpretable,lundberg2017unified}. Examples of perturbation-based approaches are LIME \cite{ribeiro2016should} and SHAP \cite{lundberg2017unified}. The main drawbacks of perturbation-based approaches are (a) that perturbed samples might not be part of the original data distribution \cite{ribeiro2018anchors}, and (b) that they are computationally inefficient, as hundreds to thousands of model evaluations are required to sample the space of local perturbations. Perturbation-based approaches are applicable to any machine-learning model \cite{ribeiro2016should,lundberg2017unified}.

\paragraph{Gradient-based Methods.} Gradient-based approaches are built on the idea of following the gradient from the output nodes of a neural network to the input nodes to obtain the features that the output was most sensitive to \cite{baehrens2010explain,simonyan2013deep}. Gradient-based approaches are therefore only applicable to differentiable models. Several improvements to this technique have since been proposed \cite{zeiler2014visualizing,smilkov2017smoothgrad,sundararajan2017axiomatic}. In particular, \cite{selvaraju2016grad} introduced the DeepLIFT method that addresses the issue of saturating gradients. 

\paragraph{Attentive Models.} Attentive models have been used in various domains to improve both interpretability \cite{xu2015show,choi2016retain,schwab2017beat,schwab2019phonemd} and performance \cite{bahdanau2014neural,yang2016hierarchical}. In computer vision, related works used attention in convolutional neural networks (CNNs) that selectively focus on input data \cite{ba2014multiple} and internal convolutional filters \cite{stollenga2014deep}. However, fundamentally, na\"ive soft attention mechanisms do not provide any incentive for a neural network to yield attention factors that correlate with feature importance. When used on top of recurrent neural networks (RNNs), attention mechanisms may propagate information across time steps through the recurrent state, and are therefore not guaranteed to accurately represent importance \cite{sundararajan2017axiomatic}. 

\paragraph{Mimic Models.} Mimic models are interpretable models trained to match the output of a black-box model. Rule-based models \cite{andrews1995survey} and tree models \cite{schwab2015capturing,che2016interpretable} are examples of models that have been used as mimic models. Mimic models are not guaranteed to match the original model and may thus not be truthful to the original model.

\paragraph{Model-agnosticism.} Irrespective of the category of the approach, we would ideally want a feature importance estimation method to be model-agnostic, i.e. independent of the choice of predictive model \cite{ribeiro2016model}. The main arguments for model-agnosticism are flexibility to choose the predictive model and feature representation as necessary \cite{ribeiro2016model}. However, in practice, the generality of model-agnostic approaches comes at a considerable cost in computational performance and scalability (Table \ref{tb:comparison}). With datasets continuously growing in size and neural networks becoming the preferred choice of model in many domains, model-specific feature importance estimation methods are often the only viable choice. This is evidenced by the recent surge in works applying model-specific approaches to analyse predictive relationships in large-scale datasets \cite{esteva2017dermatologist,ilse2018attention}. In addition, it would be desirable to have a measure of the expected quality of the provided importance estimates. Against the backdrop of estimates that are potentially not truthful to the underlying data, such a measure would enable us to assess the expected estimation accuracy and inform us when accurate estimates can not be expected. To the best of our knowledge, the presented Granger-causal objective is both the first tool that quantifies the expected quality of importance estimates, and the first objective that enables neural networks to learn to estimate feature importance.

\section{Attentive Mixtures of Experts}
\label{sec:ame}

We consider the setting in which we are given a dataset containing training samples $X$. Each $X$ consists of input features $x_i$ with $i \in [1, p]$ where $p$ is the number of input features per sample. A ground truth label $y_{\text{true}}$ is available for each training sample. Using the labelled training dataset, we wish to train a model that produces (1) accurate output predictions $y$ for new samples for which we do not have labels, and (2) feature importance scores $a_i$ that correspond to the importance of each respective input feature $x_i$ towards the output $y$. To model this problem setting, we introduce AMEs, a mixture of experts model that consist of $p$ experts $E_i$ and their corresponding attentive gating networks $G_i$ (Figure \ref{fig:ame_overview}). At prediction time, the attentive gating networks output an attention factor $a_i$ for each expert to control its respective contribution $c_i$ to the AME's final prediction $y$. All of the experts and the attentive gating networks are neural networks with their own parameters and architectures\footnote{The experts are, however, not separate models because all parts of AMEs are connected, differentiable, and trained end-to-end. An AME is therefore a single model and not an ensemble of models.}. AMEs do not impose any restrictions on the experts other than that they need to expose their topmost feature representation $h_i$ and their contribution $c_i$ for a given $X$. 

As input to the gating networks, the hidden states $h_i$ and local contributions $c_i$ of each expert are concatenated to form the combined hidden state $h_\text{all}$ of the whole AME:
\begin{small}
\begin{align}
\label{eq:hc}
h_\text{all} = \text{concatenate}(h_1, c_1, h_2, c_2, ..., h_p, c_p)
\end{align}
\end{small}
We denote $c_i$ as the output $E_i(x_i)$ of the $i$th expert for the given feature of the input data $x_i$. $a_i$ represents the output $G_i(h_\text{all})$ of the $i$th attentive gating network $G_i$ with respect to the combined hidden state $h_\text{all}$ of the AME. The output $y$ of an AME is then given by:
\begin{small}
\begin{align}
\label{eq:y}
y = \sum_{i=1}^{p} \underbrace{G_i(h_\text{all})}_{a_i} \underbrace{E_i(x_i)}_{c_i}
\end{align}
\end{small}
\noindent The attention factors $a_i$ modulate the contribution $c_i$ of each expert to the final prediction $y$ based on the AME's combined hidden state $h_\text{all}$. The attention factors therefore represent the importance of each expert's contribution towards the output $y$. The motivation behind structuring AMEs as a mixture of experts with input features $x_i$ distributed across experts is to ensure (1) that each expert's contribution $c_i$ can \textit{only} be based on their respective input feature $x_i$, and (2) that the importance of $c_i$ towards the final prediction $y$ can only be increased by increasing its respective attention factor. Splitting the features across experts guarantees that there can not be any information leakage across features, and that the attention factors $a_i$ can in turn safely be interpreted as the importance of the input feature $x_i$ towards $y$ upon model convergence. We calculate the attention factors $a_i$ using:
\begin{small}
\begin{align}
\label{eq:a_i}
a_{i} &= \frac{\exp(u_{i}^Tu_{s,i})}{\sum_{j=1}^p \exp(u_{j}^Tu_{s,i})}
\end{align}
\end{small}
where
\begin{small}
\begin{align}
\label{eq:u_i}
u_{i} &= \text{activation}(W_{i}h_\text{all} + b_{i})
\end{align}
\end{small}

\noindent corresponds to a single-hidden-layer multi-layer perceptron (MLP) with an activation function, a weight matrix $W_{i}$ and bias $b_{i}$. To compute the attention factors, we first feed the combined state $h_\text{all}$ of the AME into the MLP to get $u_i$ as a projected hidden representation of $h_\text{all}$ \cite{xu2015show,rocktaschel2015reasoning,yang2016hierarchical}. We then compute the similarity of the projected hidden representation $u_i$ to a per-expert context vector $u_{s,i}$. The context vector $u_{s,i}$ can be seen as a fixed high-level representation that answers the question: "What projected hidden representation would be the most informative for this expert?". The per-expert context vector $u_{s,i}$ is initialised randomly and has to be learned with the other network parameters during training. We obtain normalised importance scores $a_i$ from the similarities through a softmax function (Eq. \ref{eq:a_i}). The attention factors $a_i$ are used to weight the contributions $c_i$ of each expert towards the final decision $y$ of the AME (Eq. \ref{eq:y}). The soft attention mechanism formulated in Equations \ref{eq:a_i} and \ref{eq:u_i} closely follows the definitions used in related works \cite{xu2015show,rocktaschel2015reasoning,yang2016hierarchical} with two notable exceptions: Firstly, we use $h_\text{all}$ rather than just the hidden representation of a single expert as input to the soft attention mechanism. This enables the AME to simultaneously take into account the information from all available experts when producing its attention factors $a_i$ and its prediction $y$, despite not sharing features in the experts themselves. Secondly, we use a separate attentive gating network for each expert to produce the attention factors. This is in contrast to existing works that use a shared representation either over feature maps in a CNN for image data \cite{xu2015show} or over the hidden states of a RNN for sequence data \cite{xu2015show,choi2016retain,yang2016hierarchical}. Using a shared or overlapping attention mechanism is problematic for importance estimation, as information from features could potentially leak across features. This is best exemplified by attention mechanisms on top of RNNs, where information can propagate across time steps through the recurrent state, and therefore influence the model output through means other than the attention factor \cite{sundararajan2017axiomatic}. The use of separate attention mechanisms prevents information leakage entirely, and at the same time enables us to apply soft attention to non-sequential and non-spatial input data.

\section{Granger-causal Objective}
\label{sec:granger_loss}
A fundamental issue of na\"ively-trained soft attention mechanisms is that they provide no incentive to learn feature representations that yield accurate attributions \cite{sundararajan2017axiomatic}. Na\"ively-trained attentive gating networks may therefore not accurately represent feature importance or even collapse towards attractive minima, such as assigning an attention weight of $a_i=1$ to a single expert and $0$ to all others \cite{bengio2015conditional,shazeer2017outrageously}. To ensure the assigned attention weights correspond to feature importance, we introduce a secondary objective function that measures the mean Granger-causal error (MGE). Granger-causality follows the Humean definition of causality that, under certain assumptions, declares a causal relationship $X \rightarrow Y$ between random variables $X$ and $Y$ if we are better able to predict $Y$ using all available information than if the information apart from $X$ had been used \cite{granger1969investigating}. Given input sample $X$, we denote $\experterror$ as the AME's prediction error without including any information from the $i$th expert and $\allerror$ as the AME's prediction error when considering all available information. To estimate $\experterror$ and $\allerror$, we use differentiable auxiliary predictors $\text{P}_{\text{aux},i}$ and $\text{P}_{\text{aux},c}$ that receive as input the concatenated hidden representations of all experts excluding the $i$th expert's hidden representation $h_\text{all}\setminus h_i$ and the concatenated hidden representations of all experts $h_\text{all}$, respectively. The auxiliary predictors are trained jointly with the AME. 
\begin{small}
\begin{align}
\label{eq:delta_eps}
y_{\text{aux},i} &= \text{P}_{\text{aux},i}(h_\text{all}\setminus h_i) \\
y_{\text{aux},c} &= \text{P}_{\text{aux},c}(h_\text{all})  
\end{align}
\end{small}
We then calculate $\experterror$ and $\allerror$ by comparing the auxiliary predictions $y_{\text{aux},i}$ and $y_{\text{aux},c}$ with the ground truth labels $y_{\text{true}}$ using the auxiliary loss function $\mathcal{L}_{\text{aux}}$. We use the mean absolute error as $\mathcal{L}_{\text{aux}}$ for regression problems and categorical cross-entropy for classification problems.
\begin{small}
\begin{align}
\label{eq:delta_eps}
\experterror &= \mathcal{L}_{\text{aux}}(y_{\text{true}}, y_{\text{aux},i})\\
\allerror &= \mathcal{L}_{\text{aux}}(y_{\text{true}}, y_{\text{aux},c}) 
\end{align}
\end{small}
Following \cite{granger1969investigating}, we define the degree $\Delta \varepsilon_i$ to which the $i$th expert is able to contribute to the final output $y$ as the decrease in error associated with adding that expert's information to the set of available information sources:
\begin{small}
\begin{align}
\label{eq:delta_eps}
\Delta \varepsilon_{X,i} &= \experterror - \allerror
\end{align}
\end{small}
This definition of $\Delta \varepsilon_{X,i}$ naturally resolves cases where combinations of features enable improvements in the prediction error - both experts would be attributed equally for the decrease. We normalise the desired attribution $\omega_i$ corresponding to the $i$th experts' attention weights $a_i$ for a given input $X$ as:
\begin{small}
\begin{align}
\label{eq:eps_norm}
\omega_i(X) = \frac{\Delta \varepsilon_{X,i}}{\sum_{j=1}^p \Delta \varepsilon_{X,j}}
\end{align}
\end{small}
Where equation (\ref{eq:eps_norm}) normalises the attributions across all experts to ensure that they are on the same scale across decisions. We calculate the Granger-causal objective $\mathcal{L}_{\text{MGE}}$ by computing the average probabilistic distance over $n$ samples between the target distribution $\Omega$, with $\Omega(i) = \omega_i$, and the actual distribution $A$, with $A(i) = a_i$, of attention values using a distance measure D. The Kullback-Leibler divergence \cite{kullback1997information} is a suitable differentiable D for attention distributions \cite{Itti_Baldi06nips}.
\begin{small}
\begin{align}
\label{eq:mge}
\mathcal{L}_{\text{MGE}} = \frac{1}{n} \sum_X \text{D}(\Omega, A)
\end{align}
\end{small}
Because the Granger-causal loss measures the average probabilistic distance of the actual attributions to the desired Granger-causal attributions, it is valid to use it as a proxy for the expected quality of explanations. A Granger-causal loss of 0 indicates a perfect match with the Granger-causal attributions. We can therefore apply the familiar framework of cross-validation and held-out test data to estimate the expected quality of the importance estimates $a_i$ on unseen data. Finally, the total loss $\mathcal{L}$ is the sum of the main loss and the Granger-causal loss weighted by a hyperparameter $\alpha$.
\begin{small}
\begin{align}
\mathcal{L} &= (1-\alpha) \mathcal{L}_{\text{main}} + \alpha\mathcal{L}_{\text{MGE}}
\end{align}
\end{small}
The core idea of the Granger-causal objective is to train predictors on distinct subsets of the input data to measure how much the exclusion of individual features reduces model performance. This approach to importance estimation is not new \cite{vstrumbelj2009explaining} and is commonly practiced in ablation studies. In addition, a similar approach, called Shapley value analysis \cite{lipovetsky2001analysis} or Shapley regression values \cite{lundberg2017unified}, has been proposed for regression using the game-theoretic concept of Shapley values \cite{shapley1953value,lundberg2017unified}. The main difference between Shapley values and Granger-causality is that feature importance in Shapley values is defined as the marginal contribution towards the model output whereas Granger-causality defines importance in terms of the marginal contribution towards the reduction in prediction error. This subtle change in definition improves computational and memory scalability from factorial to linear in the number of features as we only have to train one additional auxiliary model per feature rather than one for every possible subset of features \cite{lipovetsky2001analysis,lundberg2017unified}.

\begin{figure*}[t]
\begin{minipage}{\textwidth}
\begin{minipage}[t]{0.22\textwidth}
\includegraphics[width=20.68ex]{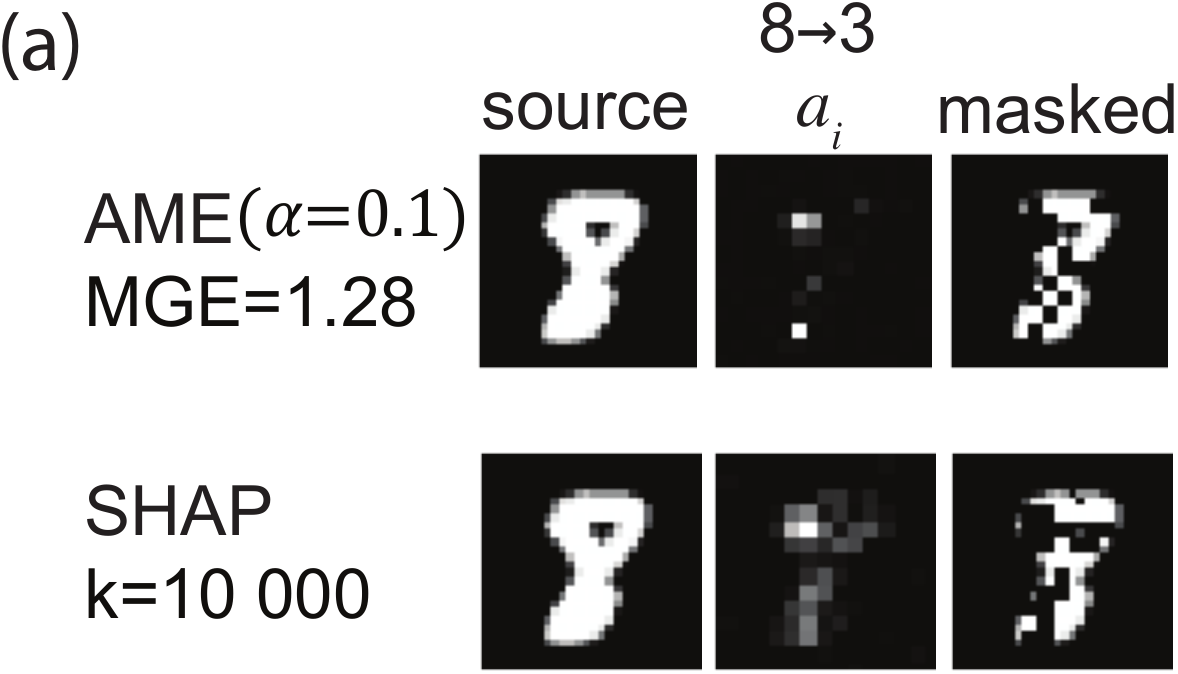}
\end{minipage}
\hskip 0.25ex
\begin{minipage}[t]{0.24\textwidth}
\includegraphics[width=22.88ex]{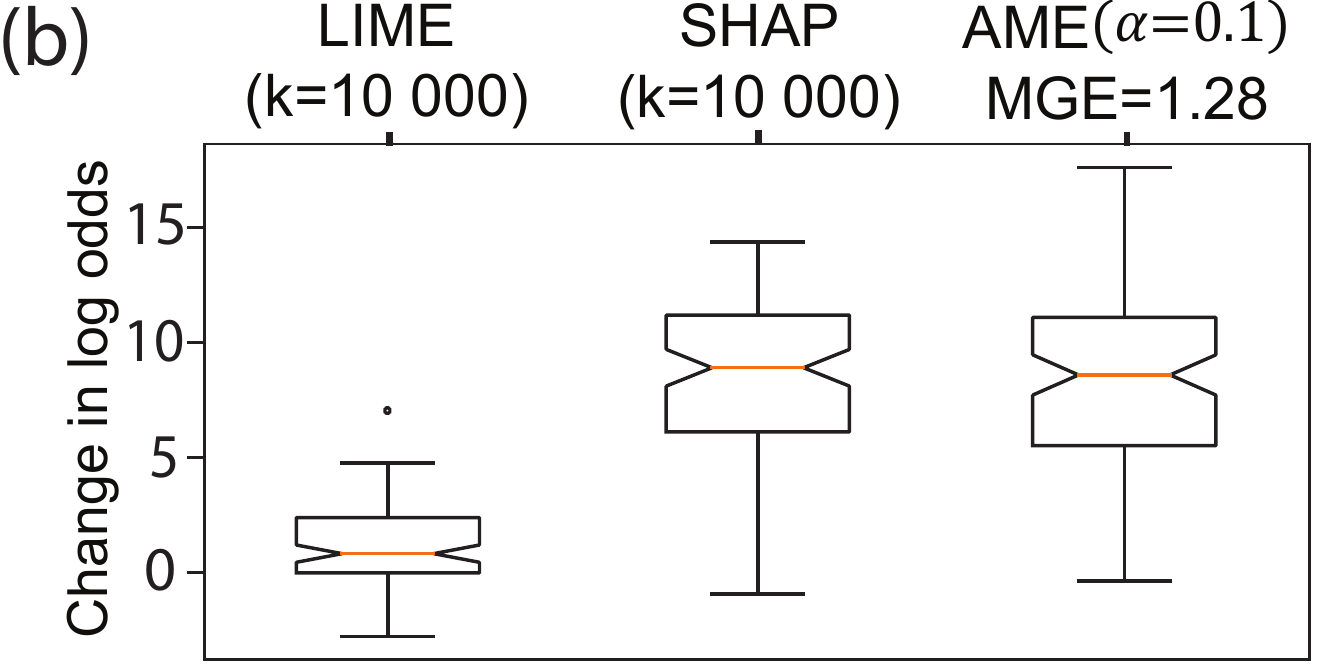}
\end{minipage}
\hskip 0.25ex
\begin{minipage}[t]{0.295\textwidth}
\includegraphics[width=29.15ex]{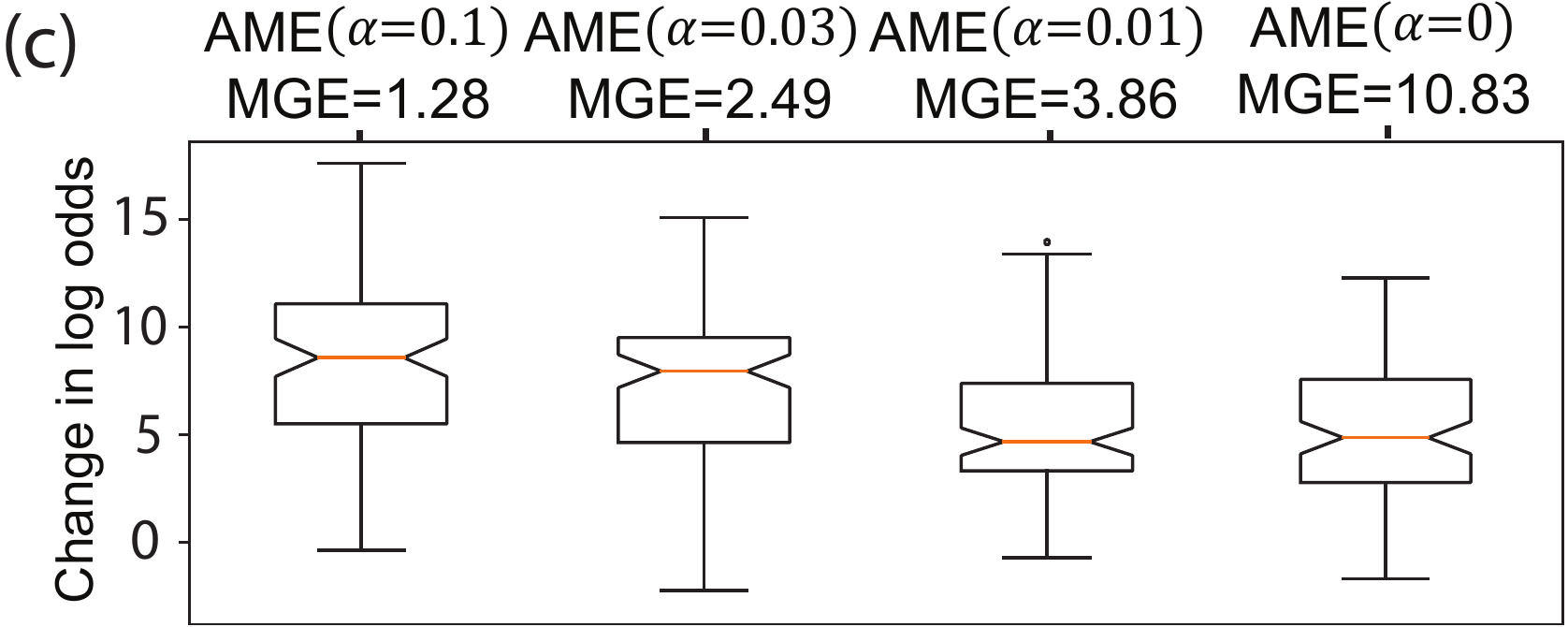}
\end{minipage}
\raisebox{11.5\height}{
\begin{minipage}[t]{0.005\textwidth}
\begin{tiny}
(d)
\end{tiny}
\end{minipage}}
\begin{minipage}[t]{0.20\textwidth}
\begin{small}
\centerline{
\begin{tabular}[b]{@{} l r @{}}
\toprule
Method & CPU(s) \\
\midrule
AME($\alpha$=$0.1$) & \textbf{3} \\
SHAP & 982 \\
LIME & 2063 \\
\bottomrule
\end{tabular}}
\end{small}
\end{minipage}
\captionof{figure}{Determining important features on MNIST. (a) The attention map $a_i$ shows which pixels were assigned the most importance. We masked the most important pixels to change the prediction to the target digit (more samples in Appendix B.2). AMEs were (b) comparable to SHAP in the change in log odds (d) at significantly lower runtime when masking over $n=100$ random images. (c) Lower MGEs correlated with better estimates when comparing AMEs with different levels of test set MGE.}
 \label{fig:8to3}
\end{minipage}
\end{figure*}

\section{Experiments}
\label{sec:evaluation}
To compare AMEs to state-of-the-art methods for importance estimation, we performed experiments on an established benchmark for importance estimation and two real-world tasks. Our goal was to answer the following questions:
\begin{itemize}[noitemsep]
\item How do AMEs compare to state-of-the-art feature importance estimation methods for neural networks in terms of estimation accuracy and computational performance?
\item Does jointly optimising AMEs for predictive performance and accurate estimation of feature importance have an adverse impact on predictive performance?
\item Does a lower test-set MGE correlate with a better expected estimation accuracy on unseen data? 
\item How do varying choices of $\alpha$ impact predictive performance and attribution accuracy?
\item Are the associations identified by AMEs and other methods consistent with those reported by domain experts? 
\end{itemize}

\subsection{Important Features in Handwritten Digits}
We performed the MNIST benchmark proposed by \cite{shrikumar2016not} to compare AMEs to LIME \cite{ribeiro2016should} and SHAP \cite{lundberg2017unified}, and to validate whether a lower test-set MGE on test data indicates a better estimation accuracy. Because AMEs provide a single set of importance scores per decision and not one set of importance scores for each possible output class, we adapted the benchmark to use a binary classifier that was trained to distinguish between a source and a target digit class ($8\rightarrow3$). We used LIME, SHAP and multiple AMEs to determine the most important pixels in an image of the source digit. The most important pixels in this setting corresponded to those pixels which distinguish the source digit from the target digit. We masked the top $10\%$ of most important pixels (Figure \ref{fig:8to3}a) and calculated the change in log odds for classifying across $n=100$ samples (Figure \ref{fig:8to3}b) to quantify to what degree the feature importance estimation methods were able to identify the important pixels for distinguishing the two digit classes \cite{shrikumar2016not,lundberg2017unified}. We brought LIME and SHAP to the same scale as the AME's attention factors $a_i$ by applying the normalising transform $a_i = \frac{\abs{e_i}}{\sum_{j=0}^n \abs{e_j} }$\inlineeqno{}. We trained AME($\alpha$=$0$) and AME($\alpha$=$0.1$) until convergence (100 epochs, 6 epochs early stopping patience) and stopped the training of AME($\alpha$=$0.03$) and AME($\alpha$=$0.01$) prematurely after 10 epochs to obtain AMEs with higher test-set MGE values for comparison (Figure \ref{fig:8to3}c). We applied LIME and SHAP to the AME($\alpha$=$0.1$) with $k$$=$$10000$ samples. Appendix B.1 lists architectures and hyperparameters. 

\subsection{Drivers of Medical Prescription Demand}
To gain a deeper understanding of what factors drive prescription demand, we trained machine-learning models to predict the next month's demand for prescription items.  

\paragraph{Dataset.} We used data related to prescription demand in England, United Kingdom during the time frame from January 2011 to December 2012. We used data streams split into six feature groups: (a) demand history, (b) online search interest, (c) regional weather, (d) regional demographics, (e) economic factors and (f) labor market data. Appendix C.1 contains a description of the dataset and the list of input features per expert (total number of features $p=585$). We applied a random split by practice to separate the data into training ($60\%$, $5673$ practices, $24.30$ million time series), validation ($20\%$, $1891$ practices, $9.07$ million time series) and test set ($20\%$, $1891$ practices, $9.07$ million time series). Because LIME and SHAP did not scale to the size of this test set, we used a subset of 3 practices ($17316$ time series) to perform the comparison on importance estimation speed.

\paragraph{Models.} We trained autoregressive integrated moving average (ARIMA) models, recurrent neural networks (RNN), feedforward neural networks (FNN), and AMEs trained with ($\alpha$\textgreater$0$) and without ($\alpha$=$0$) the Granger-causal objective. Each feature group was represented as an expert in the AMEs for a total of six expert networks. The AMEs trained without the Granger-causal objective served as a baseline of relying on neural attention only.  ARIMA served as a baseline that did not make use of any information apart from the revenue history. We applied all feature importance estimation methods except DeepLIFT to the same AME($\alpha$=$0.5$). Because there, to our knowledge, currently exists no DeepLIFT propagation rule for attentive gating networks, we used the highest-performing FNN to produce the DeepLIFT explanations (architectures in Appendix C.2).

\begin{table}[b!]
\begin{small}
\caption{Comparison of the symmetric mean absolute percentage error (SMAPE; in \%) on the test set of 1891 practices ($n = 9.07$ million time series), and the average $\pm$ standard deviation of CPU hours used for training and evaluation across the 35 runs. }
\centerline{\begin{tabular}{l@{\hskip 17ex}*{2}{r}}
\toprule
\centering
\label{tb:results}
\hspace{-0.7ex}Method & SMAPE (\%) & CPU (hr)\\
\midrule
RNN & \textbf{32.79} & 0.25$\pm$0.07  \\ 
FNN & 32.87 &  \textbf{0.06}$\bm\pm$\textbf{0.02}  \\ 
AME($\alpha$=$0$) & 33.08 & 0.45$\pm$0.14   \\ 
AME($\alpha$=$0.04$)\hspace{2ex} & 33.85 & 0.21$\pm$0.08   \\ 
ARIMA & 34.98& 527.96 \\
\bottomrule
\end{tabular}}
\end{small}
\end{table}

\paragraph{Hyperparameters.} For all neural networks, we performed a hyperparameter search with hyperparameters chosen at random from predefined ranges ($L=1-3$ hidden layers, $N=8-128$ hidden units per layer, $0-80\%$ dropout) over 35 runs. We selected those models from the hyperparameter search that achieved the best performance. Methodologically, we optimised the neural networks' mean squared error (MSE), batch size of 256, with an early stopping patience of 12 epochs and a learning rate of $0.0001$. For ARIMA, we used the iterative parameter selection algorithm from \cite{hyndman2007automatic}. To better understand the impact of $\alpha$, we trained AMEs with $\alpha\in [0, 0.1]$ chosen on a grid in steps of $0.01$. We used a neighbourhood of $k=2000$ perturbed samples for LIME. Despite our use of a small subset for the comparison on estimation speed, we were only able to apply SHAP with $k=5$ perturbed samples. The expected computation time for applying SHAP with $k=100$ perturbed samples was 9 months of CPU time.

\paragraph{Pre- and Postprocessing.} Prior to fitting the models, we standardised the prescription revenue history data for each time series to the range $[0, 1]$. We normalised all other features to have zero mean and unit variance.

\paragraph{Metrics.} We compared the predictive accuracy of the different models by computing their symmetric mean absolute percentage error (SMAPE) \cite{flores1986pragmatic} on the test set of 1891 practices. We additionally compared the speed of the various feature importance estimation methods by measuring the computation time in CPU seconds used for evaluation and the time in CPU hours used for training.

\paragraph{Results.} For importance estimation, AMEs (2 CPU seconds) were faster than DeepLIFT (24 CPU seconds), LIME (10464 CPU seconds) and SHAP (729068 CPU seconds) by one, four and six orders of magnitude, respectively. In terms of predictive performance, AME($\alpha$=$0$) models performed slightly worse than the FNN and RNN (Table \ref{tb:results}). Furthermore, AME($\alpha$=$0.04$) performed worse than AME($\alpha$=$0$). This indicates that there was a small performance decrease associated with both (i) the use of attentive gating networks, and (ii) optimising jointly to maximise predictive performance and feature importance estimation accuracy. We hypothesise that (ii) is caused by adverse gradient interactions \cite{doersch2017multi,schwab2018not} between the main task and the Granger-causal objective. We also found that AMEs indeed effectively learn to match the desired Granger-causal attributions (Eq. \ref{eq:eps_norm}) with a Pearson correlation $r^2$ of $0.84$ measured on the test set. In contrast, the AME($\alpha$=$0$) trained without the Granger-causal objective only reached a $r^2$ of $0.19$. The training time of AME($\alpha = 0.04$) was comparable to RNNs. 

\begin{figure}[t!]
 \caption{The mean value (solid lines) and the standard deviation (shaded area) of the MSE (purple) and the MGE (grey) of AMEs trained with varying choices of $\alpha \in [0, 0.1]$ across 35 runs on the test set of 1891 practices.}
  \centerline{\includegraphics[width=\columnwidth]{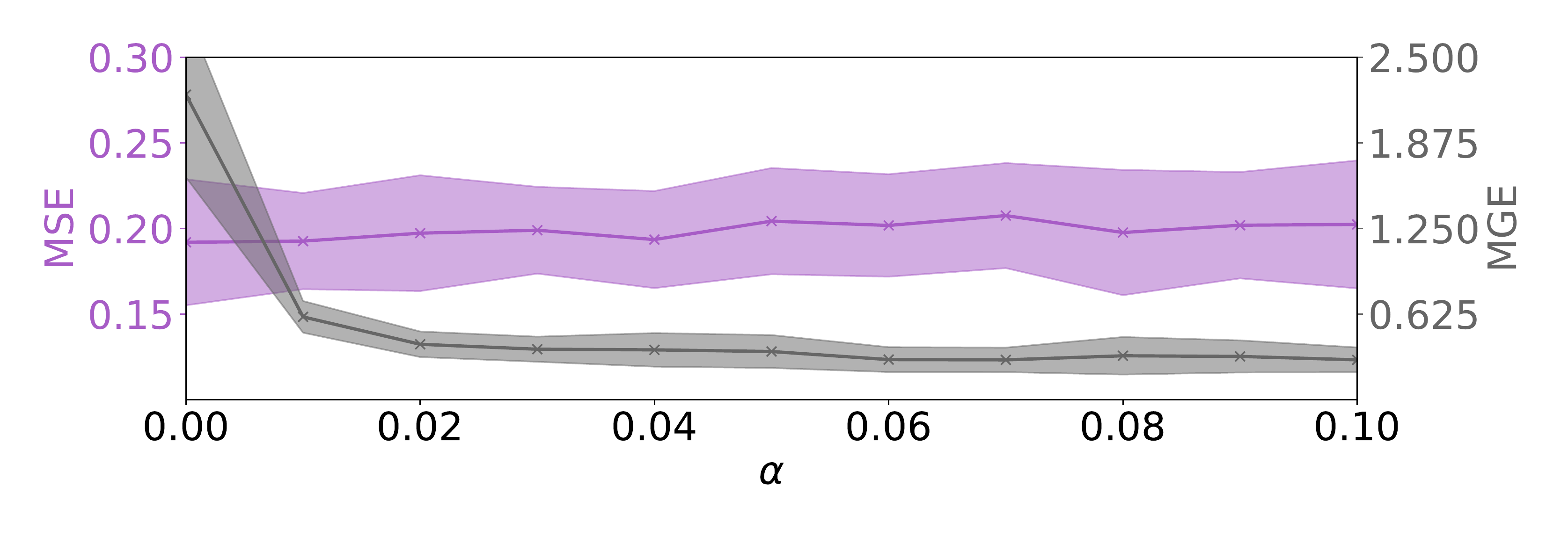}}
  \label{fig:alpha_sensitivity}
\end{figure}

\paragraph{Impact of $\alpha$.} Increasing values of $\alpha$ in the range of $[0,0.1]$ lead to an exponential improvement in MGE that was accompanied with a minor decrease in MSE (Figure \ref{fig:alpha_sensitivity}). A good middle ground was at $\alpha\approx0.03$, where roughly 80\% of the attribution accuracy gains were realised while maintaining most of the performance. The relationship between MGE and MSE was constant for values of $\alpha>0.1$.

\subsection{Discriminatory Genes Across Cancer Types}
To pinpoint the genes that differentiate between several types of cancer, we analysed the feature importances in machine-learning models trained to classify gene expression data as being either breast carcinoma (BRCA), kidney renal clear cell carcinoma (KIRC), colon adenocarcinoma (COAD), lung adenocarcinoma (LUAD) and prostate adenocarcinoma (PRAD). 

\paragraph{Dataset.} We used gene expression data from multiple cancer types in 801 individuals from The Cancer Genomic Atlas (TCGA) RNAseq dataset. To keep visualisations succinct, we used a subset of 100 genes as input data. We applied a stratified random split to separate the data into training (60\%, 480 samples), validation (20\%, 160 samples) and test set (20\%, 161 samples). 

\paragraph{Models.} We trained FNN, AME($\alpha$=$0$) and AME($\alpha$=$0.05$) (architectures in Appendix D.1). LIME($k$=$400$) and SHAP($k$=$100$) were applied to the AME($\alpha$=$0.05$) and DeepLIFT to the best FNN for the same reason as in experiment 2. We also trained five random forests (RF) \cite{breiman2001random} with $2048$ trees in a binary one-vs.-all classification setting for each cancer type. We used the Gini importance measure \cite{breiman2001random,genuer2010variable,louppe2013understanding} derived from the RFs as a baseline that was independent of the neural networks.

\paragraph{Hyperparameters.} For each of the 100 gene loci, we used a MLP with a single hidden layer with batch normalisation \cite{ioffe2015batch} and a single neuron as expert networks in the AME models. Each expert network received the gene expression at one gene locus as its input. For the FNN baseline, we chose the matching hyperparameters and architecture (100 neurons, 1 hidden layer). We optimised the neural networks with a learning rate of 0.0001, a batch size of 8 and an early stopping patience of 12 epochs. We trained each model on 35 random initialisations.
\begin{table}[t!]
\begin{small}
\caption{Comparison of the number of gene-cancer associations that were substantiated by literature evidence in the top 10 genes by average importance (Recall@10), and the number of CPU seconds used to compute them.}
\centerline{
  \begin{tabular}{l@{\hskip 23.5ex}*{2}{r}}
  \toprule
Method & Recall@10 & CPU(s) \\
\midrule
AME($\alpha$=$0.05$) & \textbf{10} & \textbf{3} \\ 
RF & \textbf{10} & 12 \\
SHAP($k$=$100$) & 8 & 6119 \\
LIME($k$=$400$) & 8 & 80 \\ 
DeepLIFT & 7 & 6 \\ 
AME($\alpha$=$0$) & 2 & \textbf{3} \\ 
\bottomrule
  \end{tabular}}
  \label{tb:recall_results}
\end{small}
\end{table}
\paragraph{Pre- and Postprocessing.} We standardised the input gene expression levels to have zero mean and unit variance. We applied the normalising transform (eq. 13) to DeepLIFT, LIME, SHAP and RF.

\paragraph{Metrics.} We compared the error rates on the test set to assess predictive performance. In order to determine whether the associations identified by the various methods are consistent with those reported by domain experts, we counted the number of gene-cancer associations that were substantiated by literature evidence in the top 10 genes by average importance on the test set (Recall@10). We performed a literature search to determine which associations have previously been reported by domain experts. Appendix D.2 contains references and details of the literature search.

\begin{figure}[t!]
  \caption{The importance of specific genes (coloured bars) for distinguishing between multiple cancer types as measured by average assignment of attention factors $a_i$. We report the average attention factors over All and over the per-cancer subsets. The grey bars spanning through the subsets highlight the 10 most discriminatory genes by average attention over All. We bolded the names of those genes whose associations are substantiated by literature evidence.}
\label{fig:gene_expression}
  \centering
      \subcaptionbox{AME($\alpha$=$0.05$)}{\includegraphics[width=25.25ex]{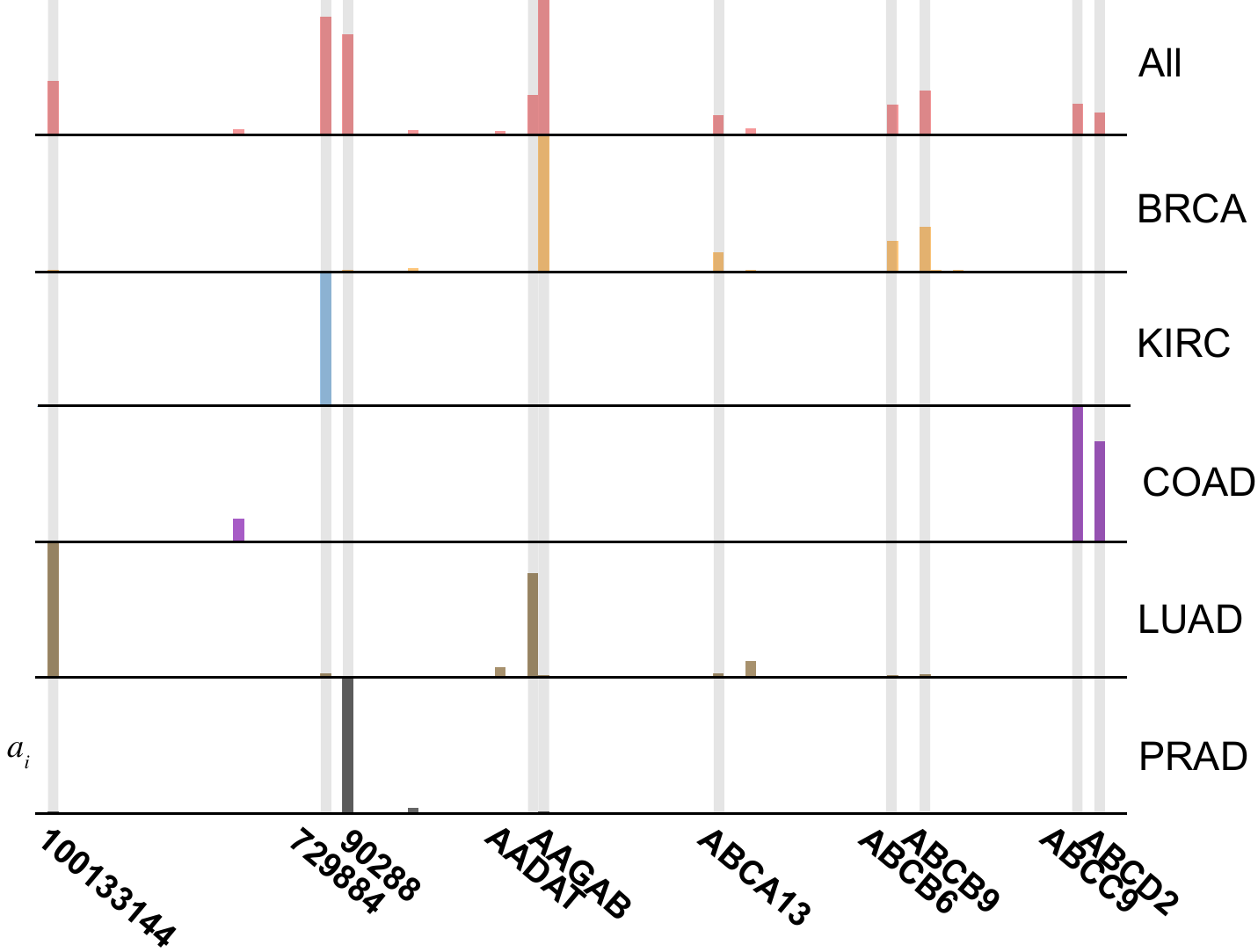}}\quad
      \subcaptionbox{AME($\alpha$=$0$)}{\includegraphics[width=25.25ex]{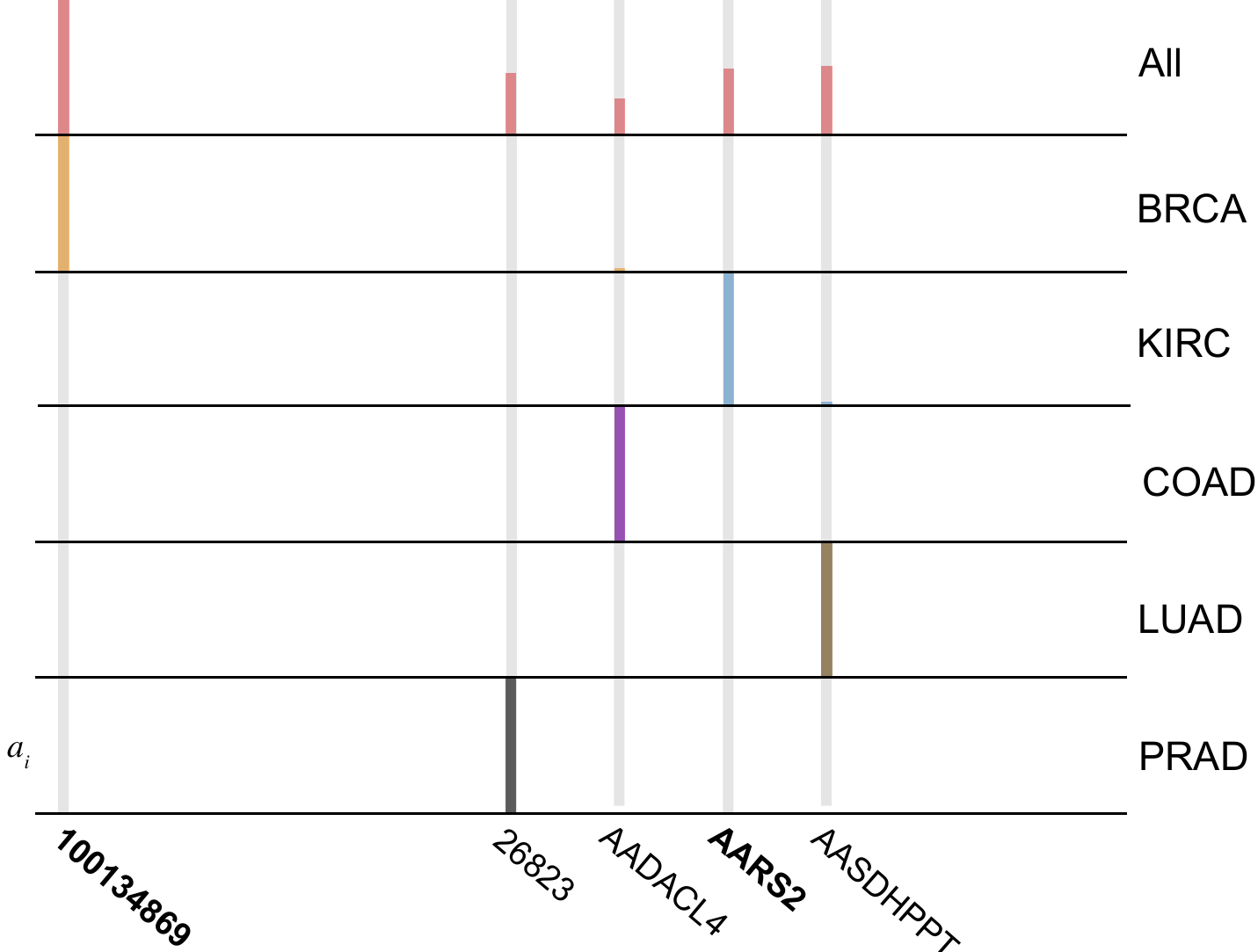}}\quad
      \subcaptionbox{DeepLIFT}{\includegraphics[width=25.25ex]{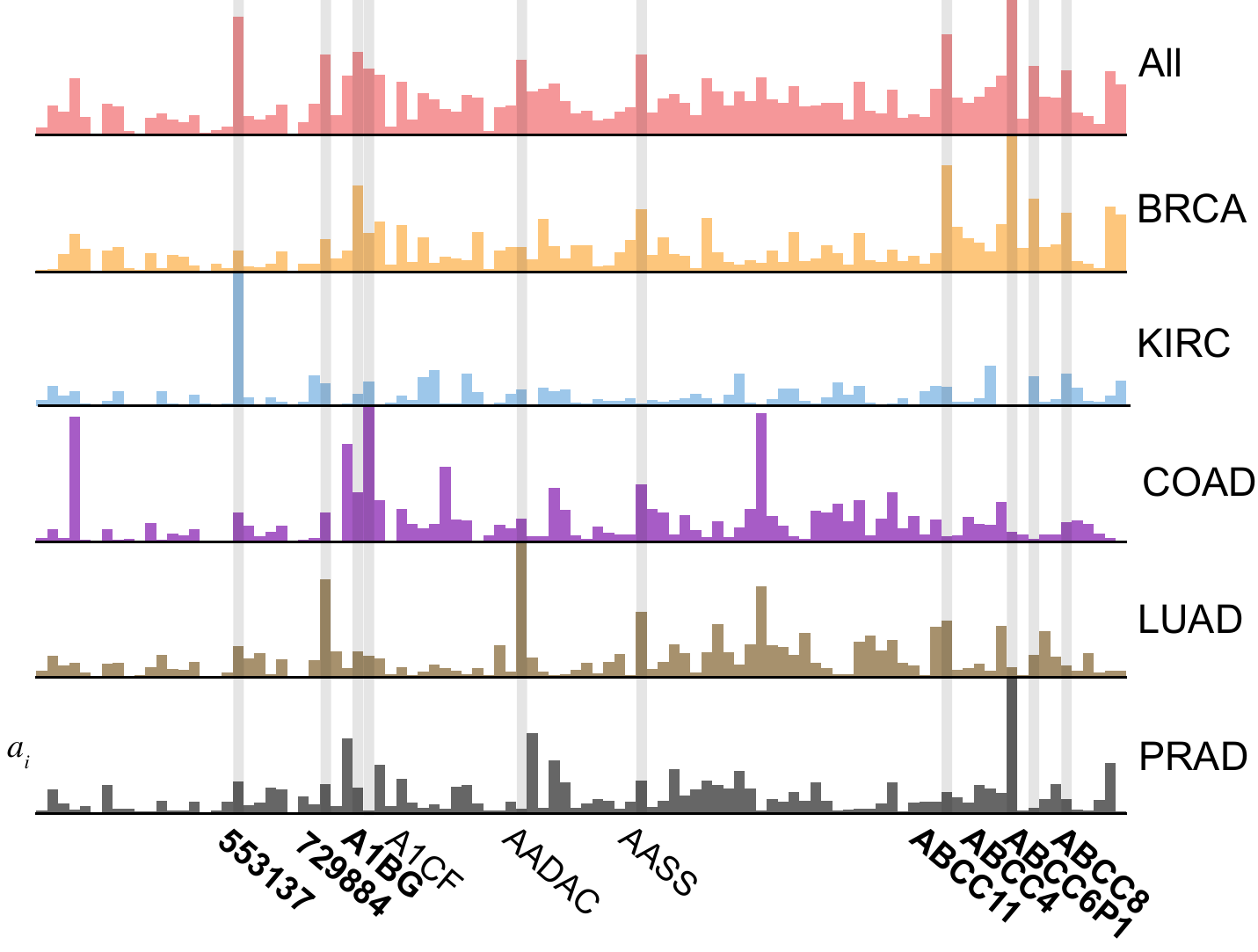}}\quad
      \subcaptionbox{LIME($k$=$400$)}{\includegraphics[width=25.25ex]{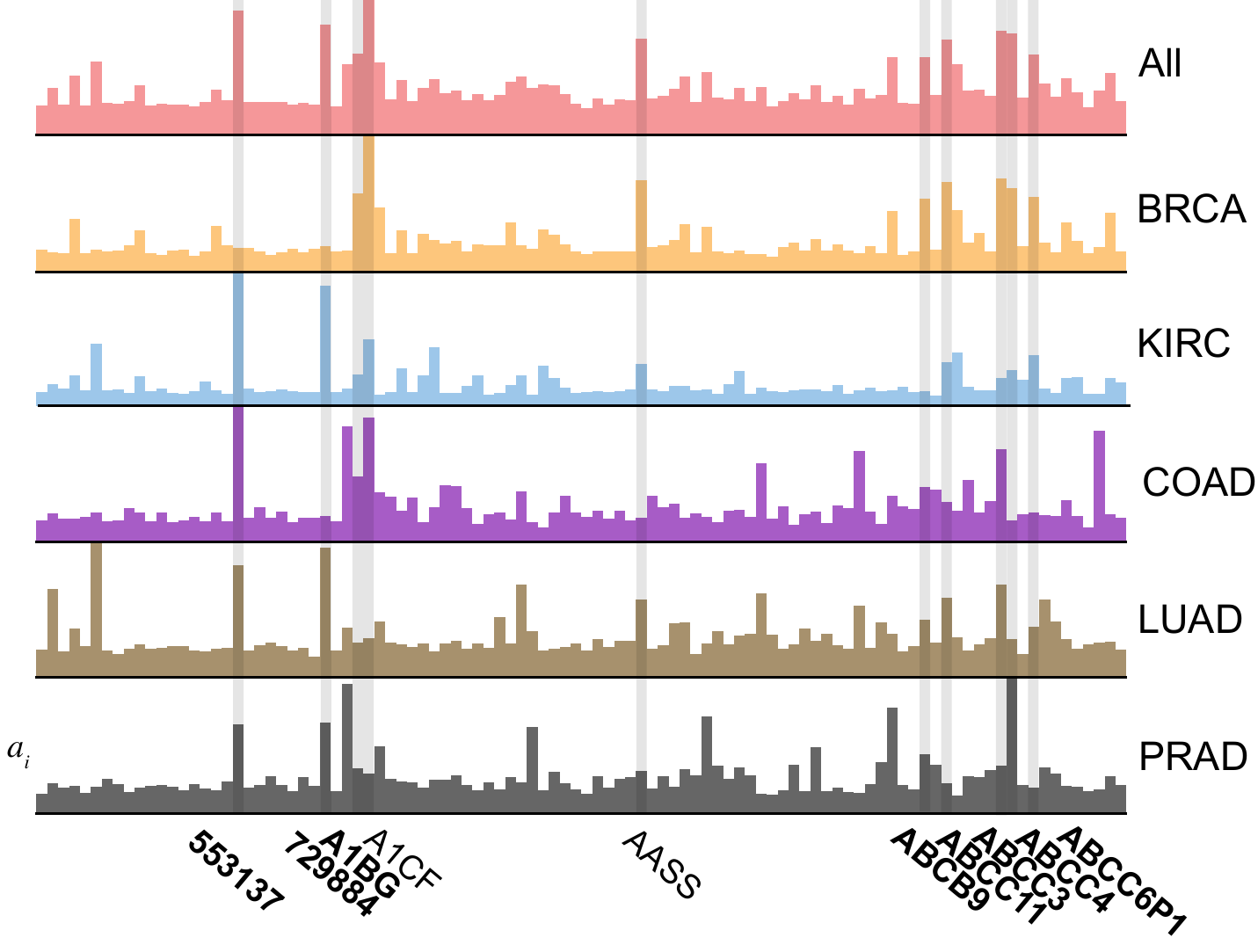}}\quad
      \subcaptionbox{SHAP($k$=$100$)}{\includegraphics[width=25.25ex]{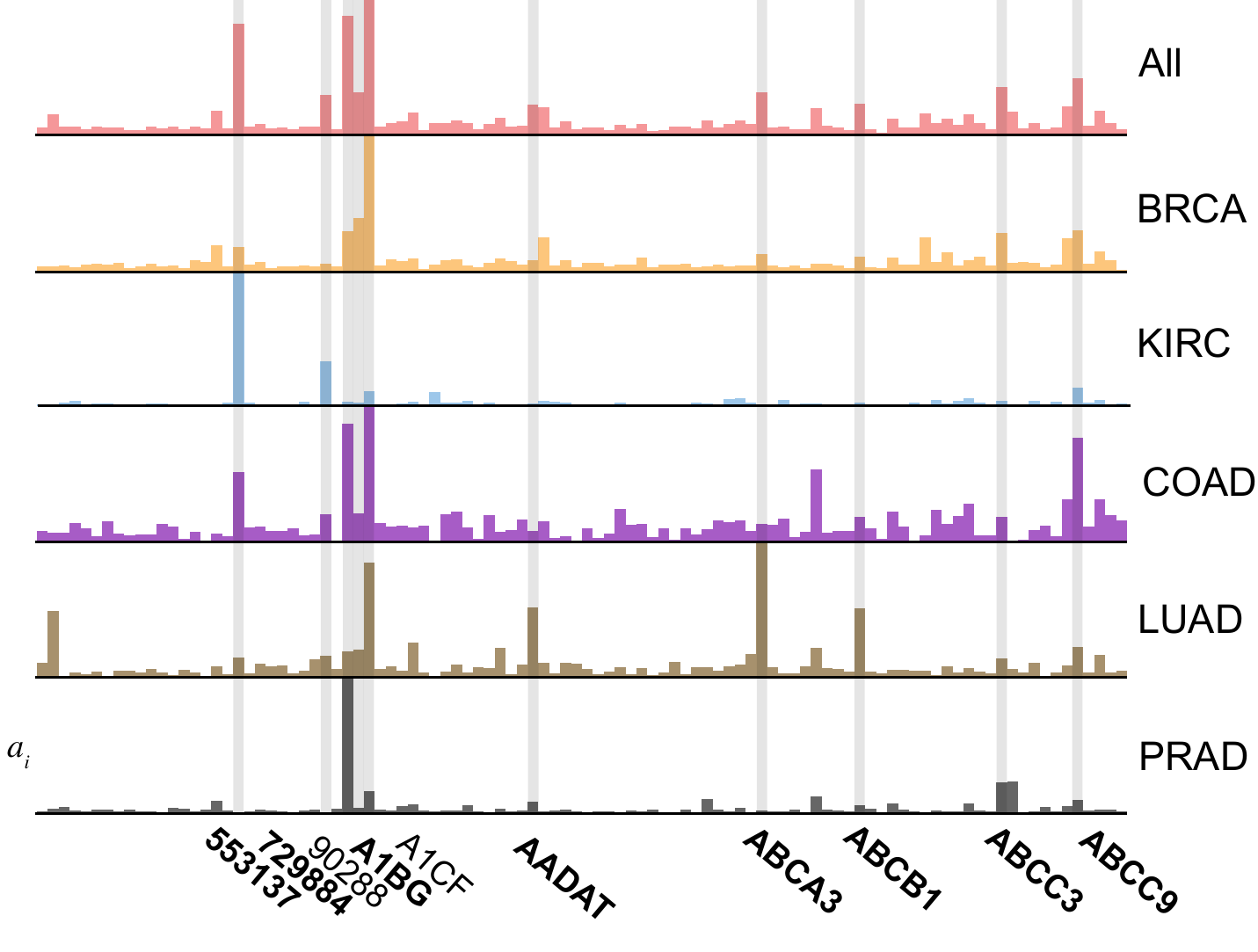}}\quad
      \subcaptionbox{RF}{\includegraphics[width=25.25ex]{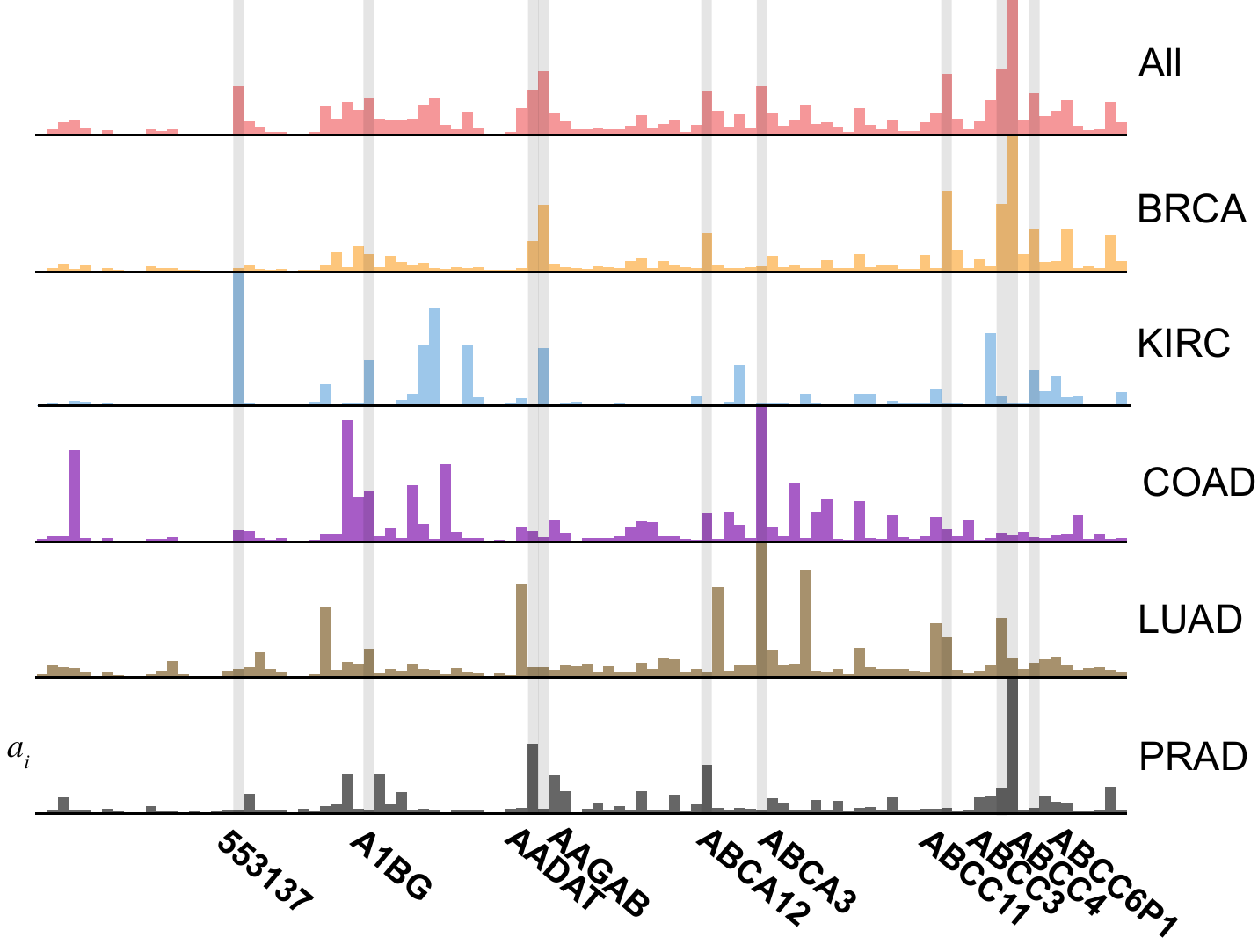}}\quad
\end{figure}

\paragraph{Results.} We found that the AME($\alpha$=$0.05$) based its decisions primarily on a small number of highly predictive genes for the different types of cancer (Figure \ref{fig:gene_expression}), and that literature evidence substantiated all of the top 10 links between respective cancer type and gene locus it reported (Table \ref{tb:recall_results}). AME($\alpha$=$0$) collapsed to assign an attention factor of 1 to one gene locus and 0 to all others for each cancer type - only reporting five non-zero importance scores. DeepLIFT, LIME and RF had difficulties discerning the important from the uninformative genes and assigned moderate levels of importance to many gene loci. DeepLIFT, LIME and SHAP were conflicted about which genes were relevant for which cancer with several of their top genes having high importance scores for multiple cancers. In contrast, AME($\alpha$=$0.05$) clearly distinguished both between cancers and important and uninformative genes. RF achieved a similar performance in terms of Recall@10 as AME($\alpha$=$0.05$). However, RFs can only produce an average set of importance scores for the whole training set. AMEs learn to accurately assign feature importance for individual samples, and can therefore explain every single prediction they make. On the test set, the mean$\pm$standard deviation of error rates across 35 runs of FNN, AME($\alpha$=$0$) and AME($\alpha$=$0.05$) were $1.10$$\pm$$0.005\%$, $4.86$$\pm$$0.093\%$, and $2.16$$\pm$$0.009\%$, respectively. 

\section{Conclusion}
We presented a new approach to estimating feature importance that is based on the idea of distributing the feature groups of interest among experts in a mixture of experts model. The mixture of experts uses attentive gates to assign attention factors to individual experts. We introduced a secondary Granger-causal objective that defines feature importance as the marginal contribution towards prediction accuracy to ensure that the assigned attention factors correlate with the importance of the experts' input features. We showed that AMEs (i) compare favourably to several state-of-the-art methods in importance estimation accuracy, (ii) are significantly faster than existing methods, and (iii) discover associations that are consistent with those reported by domain experts. In addition, we found that there was a trade-off between predictive performance and accurate importance estimation when optimising jointly for both, that training with the Granger-causal objective was crucial to obtain accurate estimates, and that a lower Granger-causal error correlated with a better expected importance estimation accuracy. AMEs are a fast and accurate alternative that may be used when model-agnostic feature importance estimation methods are prohibitively expensive to compute. We believe AMEs could therefore be a first step towards translating the strong performance of neural networks in many domains into a deeper understanding of the underlying data. 

\subsubsection*{Acknowledgments}
This work was funded by the Swiss National Science Foundation project No. 167302. We acknowledge the support of the NVIDIA Corporation. Contains public sector information licensed under the Open Government Licence, and data generated by the TCGA Research Network: \url{http://cancergenome.nih.gov/}.

\small 
\bibliographystyle{aaai}
\bibliography{references.bib}


\end{document}